%% file: root.tex
\documentclass[letterpaper, 10 pt, conference]{ieeeconf}  

\IEEEoverridecommandlockouts                              

\usepackage{color,array}
\usepackage{graphicx}

\usepackage{booktabs}
\usepackage{multirow}
\usepackage{colortbl}
\usepackage{todonotes}
\usepackage{makecell}
\usepackage{amssymb}
\usepackage{amsmath}
\usepackage{dblfloatfix} 
\usepackage{array}
\usepackage{hyperref}
\usepackage{tabularx}

\definecolor{lightgreen}{rgb}{0.88, 1, 0.88}
\definecolor{tabfirst}{rgb}{1, 0.7, 0.7} 
\definecolor{tabsecond}{rgb}{1, 0.85, 0.7} 
\definecolor{tabthird}{rgb}{1, 1, 0.7} 

\title{\LARGE \bf
SPIN: An Open Simulator of Realistic Spacecraft Navigation Imagery
}

\author{Javier Montalvo$^{*1}$, Juan Ignacio Bravo Pérez-Villar$^{*}$, Álvaro García-Martín, \\ Pablo Carballeira, Jesús Bescós 
\thanks{*These authors contributed equally}
\thanks{$^1$ Corresponding author, email: javier.montalvor@estudiante.uam.es}%
\thanks{All authors are with the VPULab from the Department of Electronics and Communications Technology at the Universidad Autónoma de Madrid.}%
\thanks{This work has been supported by the Ministerio de Ciencia, Innovación y Universidades of the Spanish Government under HVD (PID2021-125051OB-I00) and SEGA-CV (TED2021-131643A-I00) projects}%
}%

\begin{document}

\maketitle
\thispagestyle{empty}
\pagestyle{empty}

\input{sec/0_abstract}

\input{sec/1_intro}

\input{sec/2_related-work}

\input{sec/3_spin-tool}
\input{sec/4_spin_validation}

\input{sec/6_conclusions}




\bibliographystyle{IEEEtran}
\bibliography{bibliography}

\end{document}

%% file: sec/0_abstract.tex
\begin{abstract}
The scarcity of data acquired under actual space operational conditions poses a significant challenge for developing learning-based visual navigation algorithms crucial for autonomous spacecraft navigation. This data shortage is primarily due to the prohibitive costs and inherent complexities of space operations. While existing datasets, predominantly relying on computer-simulated data, have partially addressed this gap, they present notable limitations. Firstly, these datasets often utilize proprietary image generation tools, restricting the evaluation of navigation methods in novel, unseen scenarios. Secondly, they provide limited ground-truth data, typically focusing solely on the spacecraft's translation and rotation relative to the camera. To address these limitations, we present SPIN (SPacecraft Imagery for Navigation), an open-source spacecraft image generation tool designed to support a wide range of visual navigation scenarios in space, with a particular focus on relative navigation tasks. SPIN provides multiple modalities of ground-truth data and allows researchers to employ custom 3D models of satellites, define specific camera-relative poses, and adjust settings such as camera parameters or environmental illumination conditions. We also propose a method for exploiting our tool as a data augmentation module. We validate our tool on the spacecraft pose estimation task by training with a SPIN-generated replica of SPEED+\cite{park2022speed+}, reaching a 47\% average error reduction on SPEED+ testbed data (that simulates realistic space conditions), further reducing it to a 60\% error reduction when using SPIN as a data augmentation method. Both the SPIN tool (and source code) and our SPIN-generated version of SPEED+ will be publicly released upon paper acceptance on GitHub. \href{https://github.com/vpulab/SPIN}{https://github.com/vpulab/SPIN}.
\end{abstract}

%% file: sec/1_intro.tex
\section{Introduction}\label{sec:introduction}

Acquiring data from space presents certain difficulties which include, but are not limited to, the high costs of designing and launching a spacecraft, the limited on-board resources for data storage and transmission, and the required precise spacecraft control to obtain the desired representative data. This limits the availability of data acquired in real-space operational conditions. While substantial efforts have been made to overcome these limitations for observation missions of the Earth~\cite{mathieu2018earth}, Sun~\cite{muller2020solar}, and other celestial bodies with high scientific value~\cite{matson2002cassini,bolton2017juno,stern2018horizon}, there is still a noticeable gap in other space-related fields. Specifically, data to support autonomous spacecraft relative navigation remains scarce~\cite{song2022deep}. 

A particular case of interest in spacecraft relative navigation is the interaction between two non-cooperative spacecrafts. This is crucial for supporting current and future space missions, including tasks such as on-orbit servicing, active debris removal, close formation flying, rendezvous and docking, or space exploration. In all these mission scenarios, autonomy is indispensable as both the signal delays and the limited bandwidth render remote spacecraft operation unfeasible.

\begin{figure}[t]
    \centering
    \includegraphics[width=\columnwidth]{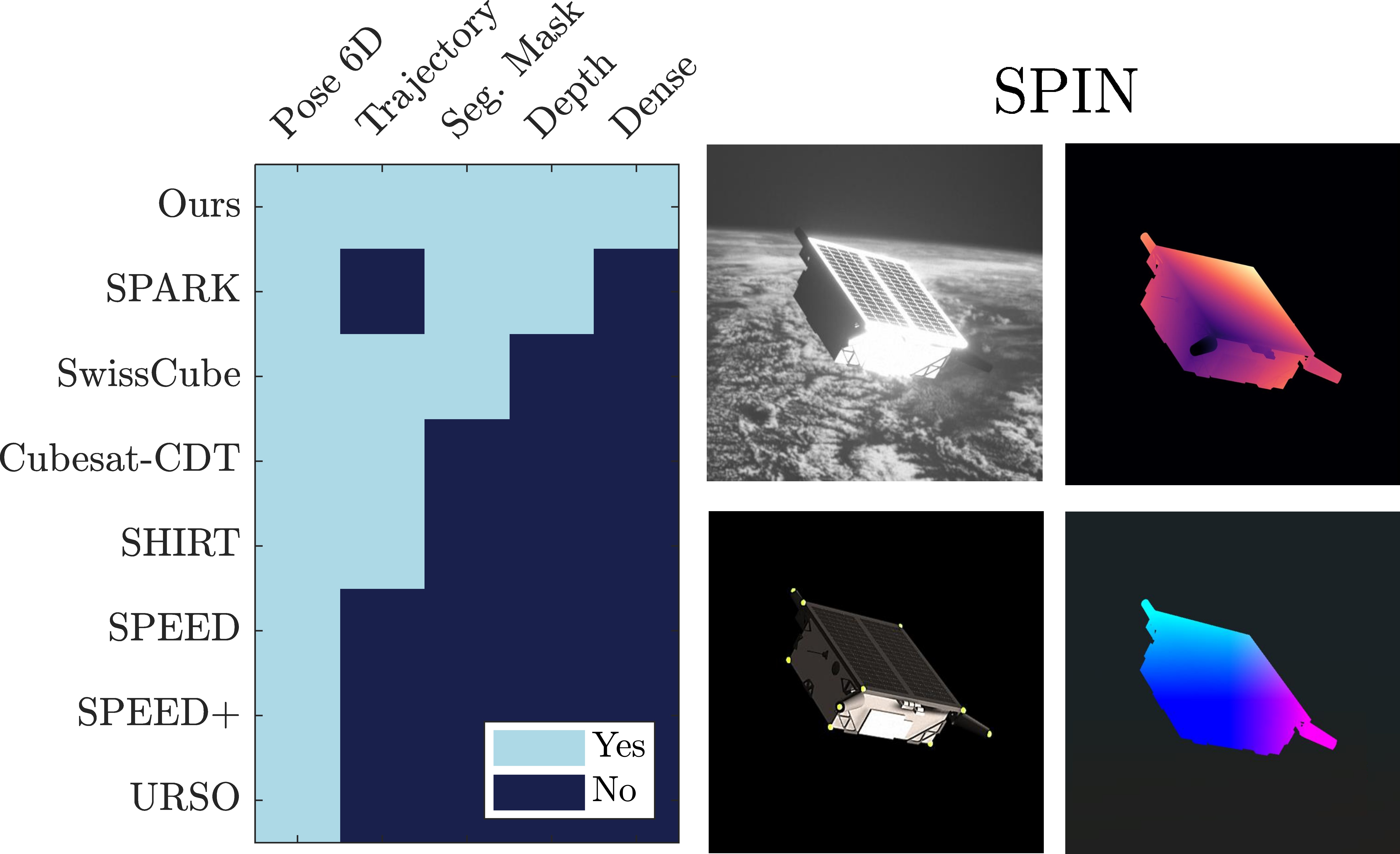}
    \caption{A comparative between SPIN and other satellite datasets. On the left, the matrix compares their features (light blue cells indicate available features). Images show SPIN data examples: a rendering and a depth ground-truth (top images); our keypoint tool and a dense-pose ground-truth (bottom images).}
    \label{fig:introduction}
\end{figure}

Visual-based navigation plays a crucial role in achieving autonomy, as it provides information on the elements of the environment and the relative position and orientation of the spacecraft. Current state-of-the-art methods for visual-based navigation employ learning techniques. However, the aforementioned lack of data prevents training robust learning-based navigation algorithms to support the required autonomy. To address this issue, researchers often turn to computer-based simulators that generate synthetic datasets, as detailed in Section \ref{sec:related-work}. These datasets are essential for research in visual-based spacecraft navigation: they have enabled the development of algorithms capable of estimating spacecraft pose with centimetre-level accuracy \cite{park2022speed+} even in situations with large differences in scale \cite{hu2021wide}.

The tools and models used for synthetic data generation are proprietary, which poses two main constraints. First, there is a difficulty in adding ground-truth data to the existing datasets which typically lack depth information, segmentation information or dense pose annotations, i.e. mapping each pixel to its corresponding location on the 3D model. Second, these proprietary tools restrict the possibility to test models in scenarios not covered by the datasets. For instance, the datasets generally do not include videos, defined here as sets of temporally related images, even though such image sequences are frequent in real rendezvous scenarios. Finally, assessing models under specific conditions like reflective surfaces or varied backgrounds, which are vital for creating more realistic images, is also challenging due to these limitations.

To bridge this gap, we present the SPacecraft Imagery for Navigation (SPIN) tool, the first open-source image generation tool designed specifically to create data for visual-based navigation between two spacecrafts. SPIN addresses the limitations of current state-of-the-art datasets, as illustrated in Fig.~\ref{fig:introduction}, by enabling the generation of highly realistic and customizable spacecraft imagery in various poses or pose sequences. It also offers extensive ground truth data, encompassing depth, dense pose annotations, segmentation, and keypoints. SPIN allows to load any external spacecraft 3D model and comes pre-loaded with a model of the Tango spacecraft.

We validate SPIN performing experiments on the task of pose estimation, the common benchmark across all identified datasets (see Fig.~\ref{fig:introduction}). We compare the results obtained from training with the SPIN-generated images against those obtained the widely used dataset SPEED+~\cite{park2022speed+}, and evaluate on the testbed images --realistic imagery captured in laboratory conditions-- from SPEED+ and SHIRT~\cite{park2023adaptive} further introduced in Section ~\ref{sec:related-work}. Training with our SPIN-generated replica from the SPEED+ dataset, yields a 47\% average reduction in error rate for spacecraft pose estimation tasks, which is further improved to an average of 60\% error rate reduction when performing data augmentation using our simulation tool.

We summarise our \textbf{contributions} as follows:
\begin{itemize}
\item We provide the first open-source simulation tool designed streamline the generation of datasets for spacecraft imagery along with depth, segmentation and dense pose ground-truth data.
\item We propose a novel method to leverage our simulator, \textit{in-scene data augmentation}, by augmenting the actual environment and not just the 2D image.
\item We share an \textit{enhanced} version of the existing SPEED+~\cite{park2022speed+} dataset with additional depth, segmentation and dense pose labels, generated using our simulation tool.
\end{itemize}

%% file: sec/2_related-work.tex
\section{Related Work}\label{sec:related-work}

In modern vision-based algorithms that employ Convolutional Neural Networks or Transformer architectures, data serves a pivotal role in facilitating effective training and achieving optimal performance. In the space operations domain, accumulating large datasets acquired in operational conditions is impractical, due to factors such as high costs, restricted on-board resources, or constrained communication links. Two primary approaches are employed to replace real space imagery: testbed facilities and computer-based simulators. Testbed facilities are specialised lab setups designed to mimic real conditions. In the context of relative spacecraft navigation, these facilities typically feature a scaled mock-up of the target spacecraft, a motion system (e.g., a robotic arm) to simulate spacecraft dynamics, authentic camera engineering models for imaging, specialised illumination systems, and the required control and computational infrastructure. With respect to simulators, rendering tools provide computer-generated images that emulate the visual characteristics of such navigation scenarios, also featuring adjustable camera parameters, customised lighting conditions, and tailored backgrounds. Additionally, they supply precise ground-truth data for aspects such as pose, depth, segmentation, and object detection. 

There exists a trade-off among cost, flexibility, and representativeness when choosing between testbed and simulator approaches. Rendering tools offer a cost-effective and flexible solution, enabling the easy generation of diverse scenarios and backgrounds. In contrast, testbed facilities utilise real hardware and accurate mock-ups, yielding images that more closely resemble actual space-operational conditions, thereby reducing the domain gap. However, the substantial costs and specialised hardware requirements associated with testbed facilities restrict their widespread adoption in open research. Consequently, they are often reserved for secondary adaptation stages or for validation and verification purposes. 

We provide a detailed description of publicly available datasets based on monocular intensity images, summarising their features and limitations in Table \ref{tab:sota-comparison}. While our focus is on optical datasets, we acknowledge the existence of datasets derived from event sensors in the literature \cite{jawaid2023towards}.

\begin{table*}[hb]
\caption{Comparison of various established satellite pose estimation datasets with our simulation tool, SPIN. A checkmark ($\checkmark$) indicates the availability of the feature, while a dash (-) indicates its absence. }
  \centering
  \resizebox{!}{2.5cm}{%
  \setlength{\tabcolsep}{6pt}
    \begin{tabular}{@{}c|cccc|cc|ccc}
    
      & \multicolumn{4}{c|}{\textbf{Labels}} 
      & \multicolumn{2}{c|}{\textbf{Images}} 
      & \multicolumn{3}{c}{\textbf{Simulation}} \\
      \cmidrule(l){2-10} 
      
      & \textbf{Pose 6D} 
      & \textbf{Dense Coord.}
      & \textbf{Depth} 
      & \textbf{Segmentation} 
      & \textbf{Bands} 
      & \textbf{Resolution} 
      & \textbf{Range} 
      & \textbf{Trajectories}
      & \textbf{Testbed}\\
      \midrule
  
      SPARK \cite{musallam2021spacecraft} 
      & $\checkmark$ & - & $\checkmark$ & $\checkmark$ & RGB & 1024x1024 & [1.5m, 10m] & - & - \\
      \midrule

      SwissCube \cite{hu2021wide}
      & $\checkmark$ & - & - & $\checkmark$ & RGB & 1024x1024 & [0.1m, 1m] & $\checkmark$ & -  \\
      \midrule
      
      CubeSat \cite{musallam2022cubesat}  
      & $\checkmark$ & - & - & - & RGB & 1440x1080 & [0.4m, 3.8m] & $\checkmark$ & $\checkmark$ \\
      \midrule

      URSO \cite{proencca2020deep}
      & $\checkmark$ & - & - & - & RGB & 1080x960 & [10m, 40m] & - & - \\
      \midrule

      SPEED  \cite{sharma2019speed}
      & $\checkmark$ & - & - & - & Gray & 1920x1200 & [3m, 40.5m] & - & $\checkmark$ \\
      \midrule
      
      SPEED+ \cite{park2022speed+}
      & $\checkmark$ & - & - & - & Gray & 1920x1200 & $\leq$ 10m & - & $\checkmark$ \\
      \midrule

       SHIRT \cite{park2023adaptive}  
      & $\checkmark$ & - & - & - & Gray & 1920x1200 & $\leq$ 8m & $\checkmark$ & $\checkmark$ \\
      \midrule     
      
      \midrule

      \textbf{SPIN} (Ours) 
      &  $\checkmark$ & $\checkmark$ &  $\checkmark$ & $\checkmark$ 
      &  RGB/Gray & Custom & Custom & $\checkmark$ & - 
    \end{tabular}%
  }

  \label{tab:sota-comparison}
\end{table*}

\subsection{Datasets}

The SPARK dataset~\cite{musallam2021spacecraft} includes over 150,000 synthetic RGB images, along with corresponding depth and segmentation masks. It features 10 different spacecraft models obtained from NASA's 3D resources and 5 distinct debris objects. The images are rendered with the Unity framework.

The SwissCube Dataset~\cite{hu2021wide} constitutes a synthetic collection featuring 50,000 images of a 1U CubeSat model based on the SwissCube satellite. These images are organized into 500 trajectories, each comprising 100 frames, and are generated using the Mitsuba Renderer 2 framework. 

The CubeSat Cross-Domain Trajectory (CDT) dataset~\cite{musallam2022cubesat} comprises RGB images captured across multiple trajectories of a 1U CubeSat within three distinct domains. Specifically, the dataset includes two synthetic domains: the first, generated using Unity, contains 50 trajectories, and the second, created with Blender, encompasses 15 trajectories. Additionally, a testbed domain is provided, featuring 21 trajectories. 

The Unreal Rendered Spacecraft On-Orbit (URSO) Dataset~\cite{proencca2020deep} is generated using Unreal Engine 4 and features a total of 15,000 synthetic RGB images. The dataset is divided into three distinct subsets, each containing 5,000 images. One subset focuses on the Dragon spacecraft, while the remaining two subsets present varying levels of complexity for the Soyuz spacecraft.

The Spacecraft Pose Estimation Dataset (SPEED)~\cite{sharma2019speed} contains 15,000 synthetic grayscale images, in addition to 305 images captured under testbed conditions. The dataset focuses on the Tango spacecraft from the PRISMA mission. Each image is annotated for pose estimation tasks.

\begin{figure}[htbp]
    \centering
    \smallskip
    \smallskip

    \includegraphics[width=\columnwidth]{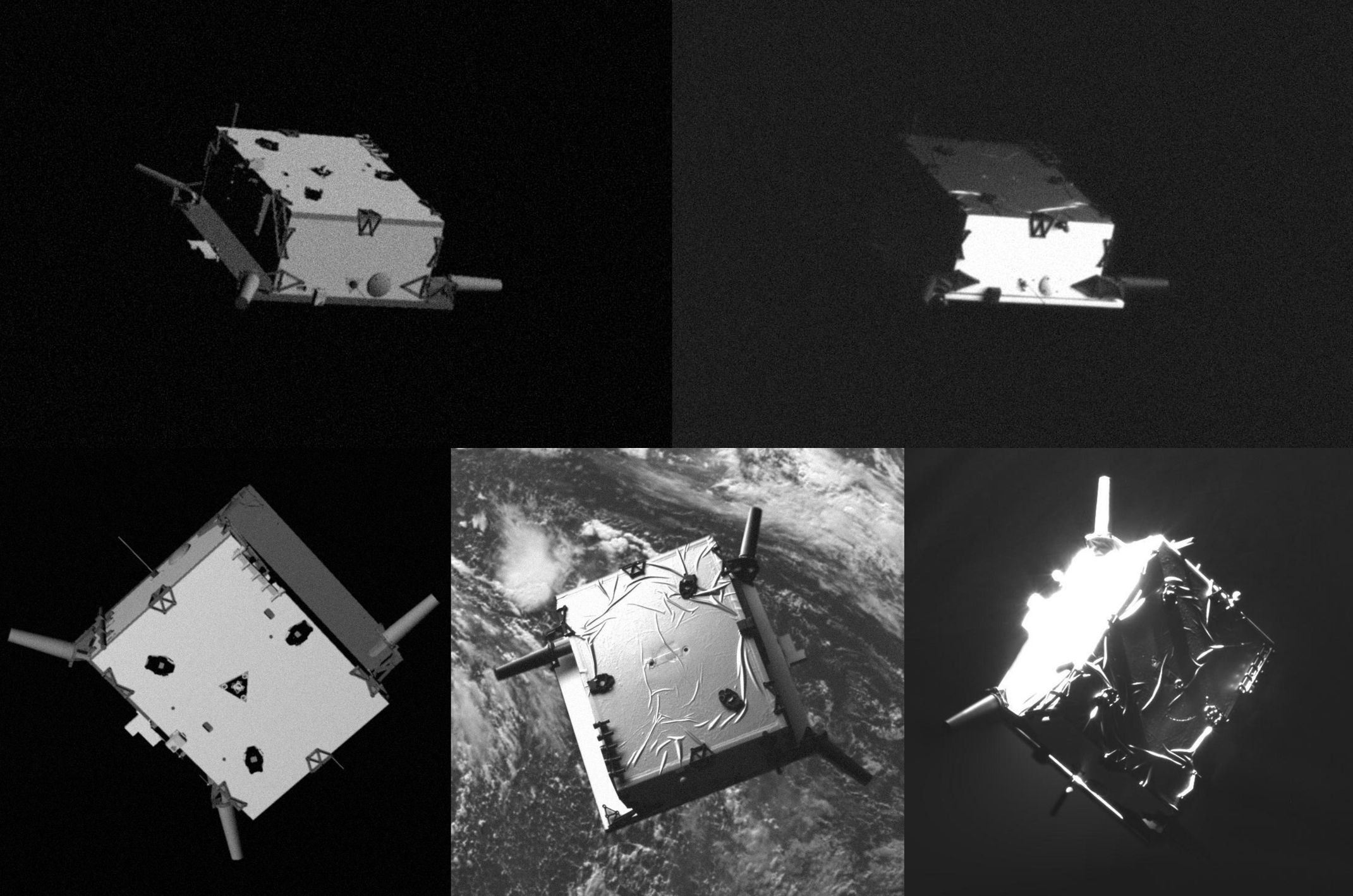}
    \caption{Representative images from the SHIRT \cite{park2023adaptive}  and SPEED+ \cite{park2022speed+} datasets. Top row  are from SHIRT dataset: Left image - synthetic domain; Right image - testbed domain. Bottom row are from SPEED+ dataset: Left to right - simulated, testbed Lightbox, testbed Sunlamp settings. }
    \label{fig:dataset-overview}

\end{figure}

The Next Generation Spacecraft Pose Estimation Dataset (SPEED+)~\cite{park2022speed+}, represented in the bottom row of Fig.~\ref{fig:dataset-overview}, consists of 60,000 synthetic images featuring the Tango spacecraft, accompanied by pose annotations. In addition, the dataset includes 9,531 annotated testbed images of a half-scale mock-up model. These test images are divided into two subsets: Sunlamp, which contains 2,791 images characterised by strong illumination and reflections against a dark background; and Lightbox, featuring 6,740 images with softer lighting conditions, elevated noise levels, and the presence of the Earth in the background.

The Satellite Hardware-In-the-loop Rendezvous Trajectories Dataset (SHIRT)~\cite{park2023adaptive}, represented in the top row of Fig.~\ref{fig:dataset-overview}, features two distinct trajectories (video sequences), ROE1 and ROE2, capturing the poses of a Tango satellite from the perspective of a service spacecraft. Each sequence offers two sets of images: synthetic grayscale images and hardware-in-the-loop testbed images that are similar to the Lightbox subset in the SPEED+ dataset.

In this work we SPEED+ and SHIRT as benchmarks due to their relevance in the spacecraft pose estimation literature. SPEED+ was employed in the European Space Agency Spacecraft Pose Estimation Challenge 2021 \cite{park2022speed+}. SHIRT expands SPEED+ to contain sequences of images.

\subsection{Discussion}

We argue that the current research in relative navigation between spacecrafts is constrained by existing datasets and the lack of simulation tools (Table \ref{tab:sota-comparison}). Firstly, datasets generally lack of diverse ground-truth labels, not providing depth, segmentation, or dense pose information. For instance, dense prediction techniques for spacecraft pose estimation have demonstrated their efficacy~\cite{ulmer20236d}, while only one dataset provides the means to compute dense pose via its dense depth maps \cite{musallam2021spacecraft}.  
Additionally, self-supervised methods for estimating monocular depth and pose~\cite{zhou2017unsupervised}, are hard to assess due to a lack of datasets with trajectories (video sequences) containing ground-truth depth. 

Secondly, the absence of image generation tools restricts the range of scenarios in which we can evaluate current state-of-the-art algorithms just to those considered by the existing datasets. Current datasets show limited variability in terms of changing surface reflectivity, camera modelling, background conditions, or variable sequences of poses. These limitations, for instance, prevent exploring research topics such as evaluating the effects of varying camera intrinsics --which could result from miscalibration during launch-- on tasks such as monocular depth estimation~\cite{Facil_2019_CVPR} or pose estimation~\cite{josephson2009pose,larsson2018camera}.

%% file: sec/3_spin-tool.tex
\begin{figure*}[!ht]
    \centering
    \includegraphics[width=\textwidth]{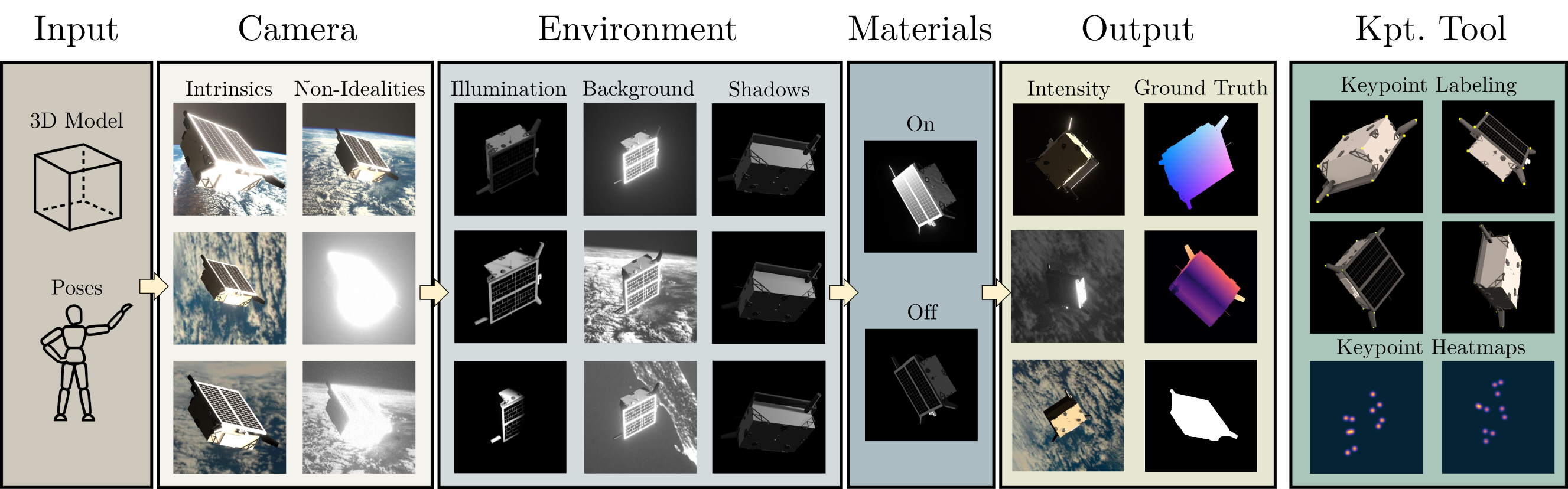}
    \caption{Pipeline of SPIN. The input to the tool is a 3D model of the spacecraft (the Tango one is the default) and a set of poses. Realism can be configured acting on three scene elements: 1) Camera, that allows to modify the intrinsics and non-idealities such as camera glare, sensor noise, and color adjustment; 2) Environment, that allows to define illumination, background, and shadows rendering; and 3) Materials, that allow enabling high-quality reflective materials. SPIN outputs the intensity images (RGB or grayscale) and ground-truth data including the dense pose, the depth, and the segmentation mask; additionally, a keypoint labeling and heatmap generation tool is provided.}
    \label{fig:tool-overview}
\end{figure*}

\section{SPIN: SPacecraft Imagery for Navigation}
Fig.~\ref{fig:tool-overview} provides a schematic overview of our proposed tool, showing its essential features, an providing an outline on different adjustable parameters from our simulator. Additionally, along the simulator we include tools to aid in the task of keypoint labeling, and also keypoint heatmap generation.

\subsection{Simulation Framework}

Our tool has been built relying on Unity Engine, which has also been previously used in related literature (CubeSat\cite{musallam2022cubesat} and SPARK \cite{musallam2021spacecraft}). Unity Engine includes multiple options for Cameras, Lightning, Materials, as well as different environment settings, and post-processing effects. 

Cameras in Unity Engine are highly configurable, and include a vast array of parameters to imitate a physical camera. From different sensor parameters such as ISO, Sensor size or pixel size, to Lens parameters such as aperture, focal length and distance. 

Unity provides different types of object-source lightning, where you can adjust some parameters such as light intensity, temperature, and whether or not the light source must cast shadows. Also, the source of light can be set up with lightning types: Directional light simulates sunlight, as if the source of light was at an infinite distance, and allows the user to set up the angular diameter for such light. Spotlight simulates a focused light aimed at a point, and user can set up the shape and reach of the spotlight, Area light simulates a soft box light , were light comes uniformly from an area and not a point in space, and the user can set up the area shape and size, and finally Point light generates light in all directions from a given point, to a maximum radius set up by the user. 

Materials in the Unity editor are customizable and allow the user to include different textures for albedo (as RGB), specular reflections, masks, and parameters such as self-emission, and textures such as normal maps and displacement maps that can be used to simulate fine detail without the computational overhead of additional polygons in the satellite model. Unity provides different shaders designed using Phisically-Based-Rendering, which we use in our simulator due to ease of use and perceived quality.

The main advantage of tools such as Unity Engine is how they enable setting up virtually any kind of environment, combining the elements previously mentioned along with 3D models, to simulate different sceneries. For this purpose, although a user may virtually be able to modify everything, it does require coding and more in-depth knowledge of specific Unity Engine libraries and elements.

In our framework, we provide an easy, straightforward way to modify these parameters and setup environments  adjusted to spacecraft imagery generation that is also easy to use and does not require specific Unity Engine knowledge or experience, so researchers that may not be familiar with 3D engines can still generate their own data adjusted to their tasks. 


\subsection{Input and Output Data}
Users can set up different environments in the simulator by setting up 3D model and poses relative to camera. By default, we include a Tango spacecraft model we built using the Speed \cite{sharma2019speed} dataset as reference, but users can also include their own 3D models and set up its materials and dimensions.

To set up camera locations, user can provide a list of relative camera poses, following the format from Speed+ \cite{park2022speed+}, describing the relative position using a quaternion $ q = {x,y,z,w}$ and a translation vector $v = {x,y,z}$.

Users can generate different modalities of data: RGB or Grayscale images, depth, dense pose or LiDAR, and ground-truth for tasks such as object detection (both 2D and 3D bounding boxes), semantic segmentation or occlusion masks. We use the Perception Unity Package \cite{unity-perception2022} to generate these multimodal data and ground-truths.

\subsection{Additional tools}
Additionally, we include a \textit{Keypoint} tool to allows users to easily extract an accuracte keypoint list for different 3D models. Related to this tool we provide another tool to generate keypoint heatmaps (methods that use keypoint heatmaps represent the state-of-the-art in spacecraft pose estimation~\cite{song2022deep}) for a given dataset using these user-generated keypoints. 

%% file: sec/4_spin_validation.tex
\section{Simulator Validation}

We propose to validate SPIN by employing its output to train a model for the task of spacecraft pose estimation, the common benchmark across the existing datasets. More in detail, we use SPIN to replicate the synthetic images of SPEED+ \cite{park2022speed+} and SHIRT \cite{park2023adaptive} for our validation (examples provided in Fig.~\ref{fig:dataset-overview}). SPEED+ shows the particularity of having a synthetic and two testbed domains which, in addition to the testbed domain of SHIRT, provide a suitable framework for evaluating the quality of SPIN and its impact on sim-to-real transfer. The experiments are organised as follows, we first describe the settings for the experiments on the pose estimation. Next, we detail the configurations used in SPIN for image generation. We conclude our validation by presenting the quantitative results, including an ablation study.

\subsection{Pose Estimation Experimental Settings} 

In all our experiments we consistently use the same architecture and the same evaluation metrics, described in Section \ref{sec:pose-estimation-model} and Section \ref{sec:metrics} respectively. All the models described are trained using PyTorch \cite{paszke2019pytorch} with input 512x512 grayscale images. We employ the Adam optimiser \cite{kingma2014adam} with a learning rate of $0.0001$. The ground-truth heatmaps are created with SPIN always using the same parameters (sigma deviation of 7 pixels) and all training parameters are kept the same for all models, including the random seed. This approach is adopted to reduce the influence of factors other than the input training data.


\subsubsection{Pose Estimation Model}\label{sec:pose-estimation-model}
We choose a simple baseline model from~\cite{xiao2018simple} to capture the performance differences introduced by the different training data. Given a spacecraft image, we use a ResNet-50-based architecture to regress a heatmap $\hat{h}~\in~\mathbb{R}^{N \times M \times C}$, where $N$ and $M$ are the image dimensions and $C$ is the number of unique keypoints. Each channel $c$ of $\hat{h}$ encodes a 2D Gaussian heatmap centered at the predicted image 2D coordinates $\hat{p}_i$ corresponding to each spacecraft 3D keypoint $P_i$. The ground-truth keypoint positions $p_i$ to generate the ground-truth heatmap $h$ are computed by projecting $P_i$ using the ground-truth pose $T$ with the perspective equation. The network is trained to minimise the mean squared error between $\hat{h}$ and $h$, as given by:
 \vspace{-2.5pt}
\begin{equation}
    \ell_{h} = \frac{1}{NMC} \sum  \Vert \hat{h} - h \rVert_F^2. \label{eq:loss_heatmap}
\end{equation}

At test time, the estimated keypoint coordinates $\hat{p}_i$ are determined by locating the maximum value in the $i^{th}$ channel of $\hat{h}$. Finally, we employ an EPnP method~\cite{lepetit2009epnp} within a RANSAC loop to retrieve the pose estimate, using the 2D-3D correspondences and the camera intrinsic parameters.

\subsubsection{Metrics}\label{sec:metrics}

We adopt the evaluation metrics defined in \cite{kelvinschallenge}. The translation error $E_v$ is calculated as the Euclidean distance between the estimated translation vector $\hat{v}$ and its ground-truth counterpart $v$, formulated as $E_v = \lVert \hat{v} - v \rVert_2$. Similarly, the orientation error $E_q$ is determined by the rotation angle required to align the estimated quaternion $\hat{q}$ with the ground-truth quaternion $q$, given by $E_q = 2 \cdot \arccos(\left| \langle \hat{q}, q \rangle \right|)$. These errors are subsequently converted into scores: the translation score is $S_v = E_v/\lVert v \rVert_2$, and the orientation score is $S_q = E_q$. Any translation and orientation scores falling below $2.173 \times 10^{-3}$ and $0.169^\circ$, respectively, are set to zero \cite{kelvinschallenge}. The total score is then computed as $S = S_q + S_v$.

\subsection{Generation Settings} 
To generate our version of the SPEED+\cite{park2022speed+} dataset using SPIN, we set some parameters to match those of SPEED+ and SHIRT\cite{park2023adaptive} with the aim of keeping a consistent input domain. Specifically, for camera settings, we keep the intrinsics, color adjustment, and noise parameters constant. This ensures that the spacecraft is viewed from the same perspective and that noise levels are uniform, for fair comparison. However, we modify glare and bloom settings to enhance the scene's realism. For the environment settings, we use shadow rendering techniques similar to those used in the SPEED+ dataset. To minimize possible effects of Earth's presence in the background for the task of pose estimation, we choose to use similar background images in our synthetic dataset as those in SPEED+. Regarding illumination, we adjust the settings to produce images with more intense directional lighting and harder shadows, resembling those that can be found on the testbed domains. 

\vspace{-5pt}
\subsection{Quantitative Results} \label{sec:results-quantitative}
To validate our dataset, we have trained the model from \cite{xiao2018simple} with our SPIN-generated dataset and with SPEED+, with the latter being used as a baseline. For both datasets, the same hyper-parameters were used, training the model for 60 epochs (120000 iterations), using a batch size of 24 images scaled to 512x512, and with a learning rate of $2.5\times10^{-4}$. 

In Table~\ref{tab:results-speed} we summarize the performance results of a model trained on SPEED+ Synthetic \textit{train} subset and a model trained on our SPIN-generated dataset with the same \textit{train} poses. Models are evaluated on both testbed subsets from SPEED+ (Lightbox and Sunlamp), and also on the testbed subsets from SHIRT (ROE1 and ROE2). We can see that the dataset generated with SPIN is able to replicate the dataset successfully. Not only that, but our more challenging training examples result in a better generalization across testbed images from both SPEED+ and SHIRT, with a 45\% reduction in the error rates for the Lightbox testbed, reducing the error from 1.764 to a 0.962, and a 64\% for the Sunlamp subset, from 2,083 to 0.760; along an error reduction of 36\% from 2.786 to 1.788 for the ROE1 testbed subset, and a 42\% for the ROE2 subset, from 2.450 to 1.421, validating our simulator  dataset generation capabilities.

\subsection{Exploiting SPIN: In-scene data augmentation}
\begin{figure}[h]
    \centering
    \includegraphics[width=.9\columnwidth]{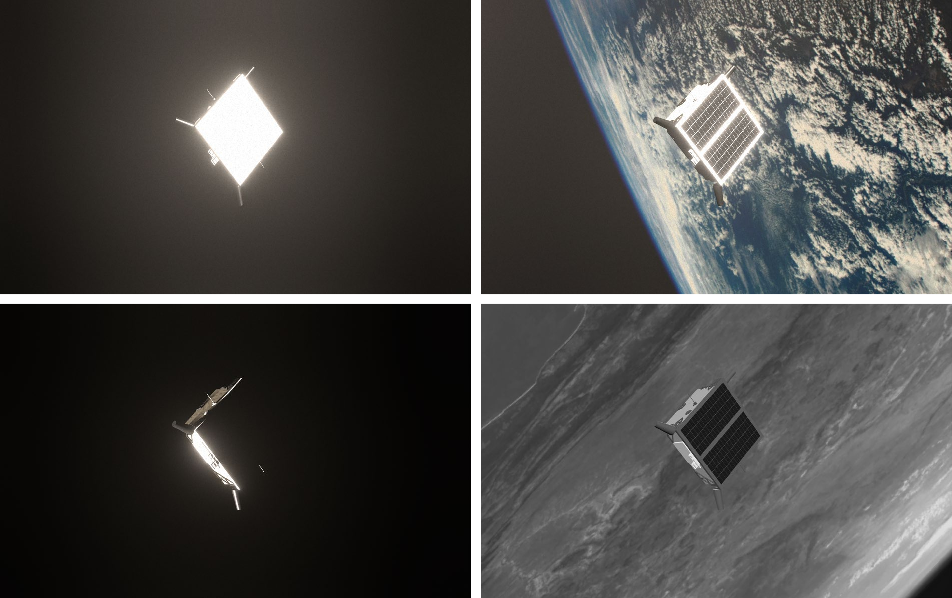}
    \caption{In-Scene Data Augmentation example. Different renderings for the same pose generated with our tool with different combinations of background, illumination, and color settings.}
    \label{fig:data-augmentation}
\end{figure}

SPIN's primary advantage over conventional datasets lies in its versatility. It enables users to generate unlimited data, opening the door to new methods and techniques that are not feasible with traditional datasets.
In computer vision, data augmentation techniques are commonly employed to significantly alter image appearances, creating diverse, unseen data to enhance deep learning model training. Our simulator achieves this goal more effectively. Instead of augmenting existing 2D images, we perform scene augmentation. This approach ensures pixel-perfect preservation of image structure while dramatically changing its appearance. For instance, we can modify illumination intensity and origin, spacecraft materials, or background. We call our proposed method \textit{in-scene data augmentation}.In Fig.~\ref{fig:data-augmentation} we show an example of the same pose augmented with our simulation tool using different rendering and lightning settings.

We evaluate the effectiveness of our method by training using three different synthetic appearances for each pose in the training dataset. We use the same hyper-parameters from our validation test, except we now train for 20 epochs to ensure that we train for the same amount of iterations (120000). Results can be found in Table~\ref{tab:results-speed}. We can see how SPIN capabilities to generate in-scene data augmentation help the model generalize better in both SPEED+ testbed subsets, with an average 21\%  error reduction, and also in SHIRT testbed subsets, with a 27\% error reduction.

\begin{table}[h]
\caption{Spacecraft pose estimation performance after training on the SPEED+ dataset and our SPIN-generated SPEED+ replicas. ISDA stands for in-scene data augmentation.}
\centering
\resizebox{!}{3.35cm}{%
\setlength{\tabcolsep}{10pt}
\begin{tabular}{@{}c|l|c|c|c}
Test Set             & Training Set     & $S_{v}\left(\downarrow\right)$        & $S_q \left(\downarrow\right)$          & $S \left(\downarrow\right)$            \\ \midrule
\multirow{3}{*}[-0.7em]{Lightbox}   & SPEED+ & 0.363          & 1.401          & 1.764         \\\cmidrule(l){2-5} 
                            & SPIN & 0.179 & 0.783 & 0.962 \\ \cmidrule(l){2-5}
                            & ISDA & \textbf{0.143} & \textbf{0.614} & \textbf{0.757} \\ \midrule
\multirow{3}{*}[-0.7em]{Sunlamp}    & SPEED+  & 0.382         & 1.701         & 2.083        \\ \cmidrule(l){2-5} 
                            & SPIN  & 0.118 & 0.642 & 0.760 \\ \cmidrule(l){2-5} 
                            & ISDA  & \textbf{0.104} & \textbf{0.490} & \textbf{0.594} \\ \midrule 
\multirow{3}{*}[-0.7em]{ROE1}       & SPEED+ & 0.665         & 2.121          & 2.786         \\ \cmidrule(l){2-5} 
                            & SPIN & 0.225          & 1.563          & 1.788\\\cmidrule(l){2-5} 
                            & ISDA & \textbf{0.155} & \textbf{1.202} & \textbf{1.357}\\\midrule 
\multirow{3}{*}[-0.7em]{ROE2}       & SPEED+ & 0.502          & 1.948          & 2.450          \\ \cmidrule(l){2-5} 
                            & SPIN & 0.283          & 1.138         & 1.421 \\\cmidrule(l){2-5} 
                            & ISDA & \textbf{0.200} & \textbf{0.797} & \textbf{0.996}     \\   \midrule

\end{tabular}%
}

\label{tab:results-speed}
\end{table}

\subsection{Ablation Test}
We conducted an ablation study to measure the impact of different SPIN settings on dataset generation and model performance. This study was motivated by our observation that the algorithm trained on our SPIN-generated replica of SPEED+ generalized better to testbed domains. We created several versions of SPEED+ using various combinations of SPIN settings. These settings were categorized into three groups: Camera settings (Glare/Bloom enabled), Environment settings (light intensity and shadows enabled), and Material settings (high-quality, reflective materials). We then measured how each setting affected the model's performance.

Table~\ref{tab:ablation-study} presents the results of the ablation study. The baseline—shown in the first row—represents the configuration with all settings disabled. We then evaluate the effect of enabling each setting individually. Activating camera effects significantly enhances performance, especially within the Sunlamp subset. While enabling environment settings also boosts performance, this improvement is observed only in the Sunlamp domain. Interestingly, activating materials alone leads to a performance drop in the Lightbox subset and offers no advantage in Sunlamp. Although the environment and material settings do not independently improve performance, their combination with camera effects shows a marked improvement, demonstrating the cumulative benefits of these settings when used together, ultimately enhancing the model trained on these dataset versions. 



 \begin{table}[h]
\caption{Impact of SPIN settings  pose estimation performance over the SPEED+ testbeds. The C,E,M letters stand for Camera, Environment and Material respectively.} 

\centering
\resizebox{0.95\columnwidth}{!}{%
\setlength{\tabcolsep}{3pt} 

\begin{tabular}{@{}ccc|ccc|ccc@{}}
\multicolumn{3}{c|}{\textbf{Settings}}     & \multicolumn{3}{c|}{\textbf{Lightbox}}           & \multicolumn{3}{c}{\textbf{Sunlamp}}             \\ \midrule
C      & E  & M   & $S_{v}\left(\downarrow\right)$        & $S_{q}\left(\downarrow\right)$        & $S\left(\downarrow\right)$            & $S_{v}\left(\downarrow\right)$        & $S_{q}\left(\downarrow\right)$        & $S\left(\downarrow\right)$            \\ \midrule
  -          & -            &  -           & 0.305          & 1.295          & 1.600          & 0.398          & 1.886          & 2.254          \\ \midrule\rowcolor{tabsecond}

$\checkmark$ & -            & -            & 0.289& 1.177& 1.466& 0.188& 1.045& 1.233\\
  -          & $\checkmark$ &  -           & 0.312& 1.293& 1.604& 0.244& 1.86& 1.430\\
   -         &      -       & $\checkmark$ & 0.395& 1.505& 1.899& 0.468& 1.767& 2.236\\ \midrule
\rowcolor{tabsecond}
$\checkmark$ & $\checkmark$ &   -          & 0.224& 0.976& 1.200& 0.139& 0.760& 0.899\\              
     -       & $\checkmark$ & $\checkmark$ & 0.344& 1.527& 1.871& 0.394& 1.711& 2.105\\
$\checkmark$ &      -       & $\checkmark$ & 0.254& 1.060& 1.314& 0.204& 1.040& 1.244\\\midrule
\rowcolor{tabfirst}
$\checkmark$ & $\checkmark$ & $\checkmark$ & 0.179& 0.783& 0.962& 0.118& 0.642& 0.760\end{tabular}%
}

\label{tab:ablation-study}
\end{table}

%% file: sec/6_conclusions.tex
\section{Conclusions}
In conclusion, SPIN offers a new tool to generate datasets with improved visuals, narrowing the gap between synthetic and real imagery in pose estimation compared to existing synthetic datasets. It also provides a wider range of ground-truth data, including the use of dense pose labels, that no other dataset in the literature provides. Moreover, SPIN facilitates the creation of new and diverse test scenarios, expanding the variety and depth of ground-truth data in current datasets.
